\documentclass{article}

\usepackage{microtype}
\usepackage{graphicx}
\usepackage{subfigure}
\usepackage{booktabs} 
\usepackage{multirow}
\usepackage{hyperref}

\usepackage[accepted]{icml2025}

\usepackage{amsmath}
\usepackage{amssymb}
\usepackage{mathtools}
\usepackage{amsthm}

\usepackage[capitalize,noabbrev]{cleveref}

\theoremstyle{plain}

\theoremstyle{definition}

\theoremstyle{remark}

\usepackage[textsize=tiny]{todonotes}

\icmltitlerunning{ID-Cloak: Crafting Identity-Specific Cloaks Against Personalized Text-to-Image Generation}

\begin{document}

\twocolumn[
\icmltitle{ID-Cloak: Crafting Identity-Specific Cloaks Against Personalized Text-to-Image Generation}



\icmlsetsymbol{equal}{*}
\icmlsetsymbol{corr}{\#}

\begin{icmlauthorlist}
\icmlauthor{Qianrui Teng}{bupt,equal}
\icmlauthor{Xing Cui}{bupt,equal}
\icmlauthor{Xuannan Liu}{bupt,equal}
\icmlauthor{Peipei Li}{bupt,corr}
\icmlauthor{Zekun Li}{ucsb}
\icmlauthor{Huaibo Huang}{cas}
\icmlauthor{Ran He}{cas}
\end{icmlauthorlist}

\icmlaffiliation{bupt}{Beijing University of Posts and Telecommunications}
\icmlaffiliation{cas}{MAIS \& NLPR, Institute of Automation, Chinese Academy of Sciences}
\icmlaffiliation{ucsb}{University of California, Santa Barbara}

\icmlcorrespondingauthor{Peipei Li}{lipeipei@bupt.edu.cn}

\icmlkeywords{Machine Learning, ICML}

\vskip 0.3in
]



\printAffiliationsAndNotice{\icmlEqualContribution} 

\begin{abstract}
Personalized text-to-image models allow users to generate images of new concepts from several reference photos, thereby leading to critical concerns regarding civil privacy. 
Although several anti-personalization techniques have been developed, these methods typically assume that defenders can afford to design a privacy cloak corresponding to each specific image. However, due to extensive personal images shared online, image-specific methods are limited by real-world practical applications. 
To address this issue, we are the first to investigate the creation of identity-specific cloaks (ID-Cloak) that safeguard all images belong to a specific identity.
Specifically, we first model an identity subspace that preserves personal commonalities and learns diverse contexts to capture the image distribution to be protected.
Then, we craft identity-specific cloaks with the proposed novel objective that encourages the cloak to guide the model away from its normal output within the subspace.
Extensive experiments show that the generated universal cloak can effectively protect the images. 
We believe our method, along with the proposed identity-specific cloak setting, 
marks a notable advance in realistic privacy protection.

\end{abstract}

\section{Introduction}

\begin{figure*}
    \centering
    \includegraphics[width=0.8\linewidth]{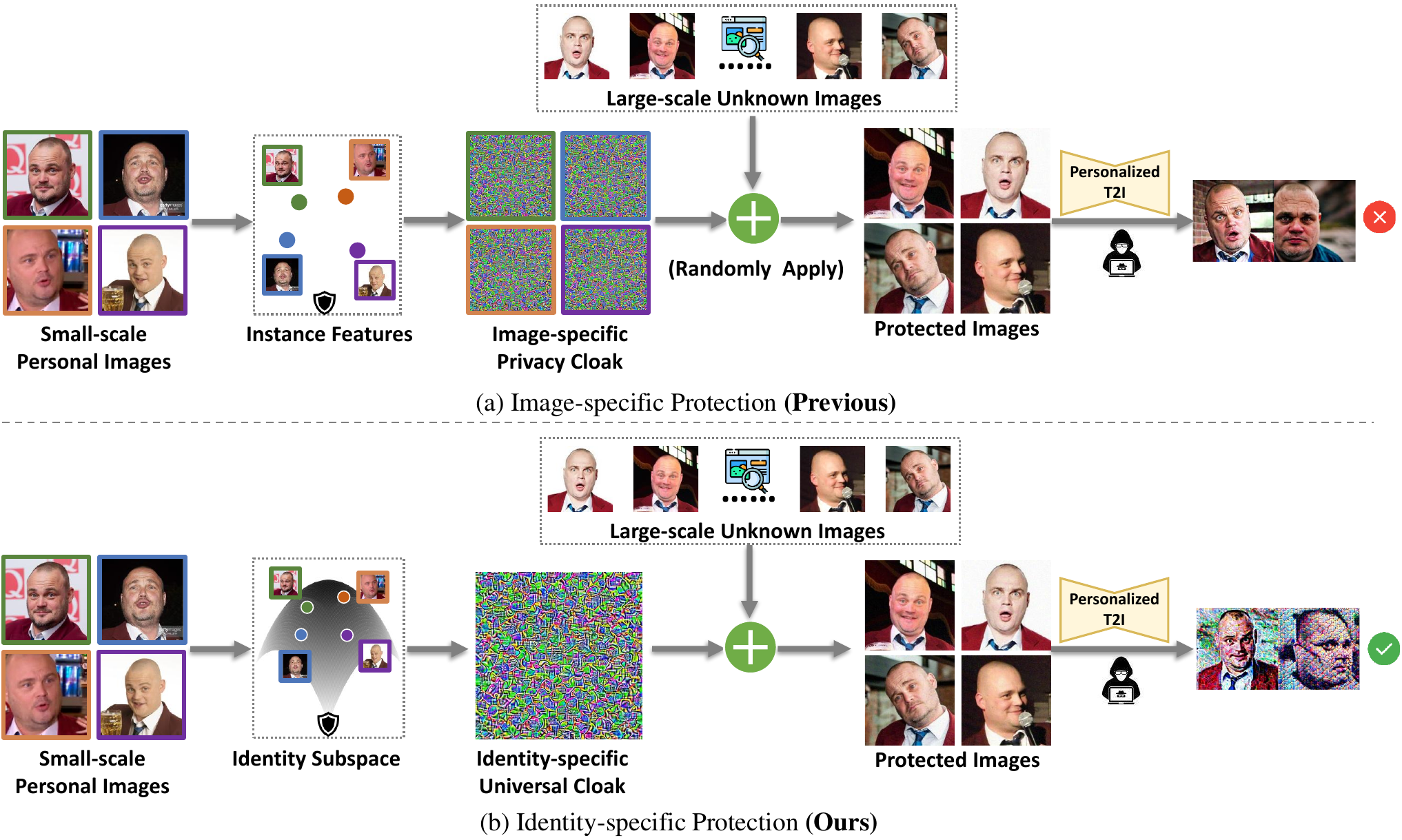}
    \vspace{-4mm}
    \caption{Comparison between image-specific and identity-specific protection. (a) Existing image-specific protection methods lose effectiveness when applied to large-scale unknown images. (b) Our identity-specific cloak exhibits effective and consistent protection.
    }
    \label{fig:intro}
\end{figure*}

With the advent of Diffusion Models~\cite{ho2020denoising, sohl2015deep, rombach2022high},  personalized text-to-image (T2I) generation~\cite{ruiz2023dreambooth,gal2022image,hu2021lora}  has ushered a novel image generation paradigm,  which enables learning new concepts to generate novel images in various contexts.
However, malicious users can easily collect personal images from social media and generate offensive fabricated images.
The potential privacy violations and the risk of image-based fraud have raised significant public concern~\cite{juefei2022countering}. Developing robust algorithms to safeguard against the malicious exploitation of diffusion models is imperative for both the research community and society.

To this end,  recent studies~\cite{liang2023adversarial,van2023anti,xue2023toward, wan2024prompt} delve into research on anti-personalization by introducing imperceptible perturbations, i.e. cloak, onto input images. Malicious users can't generate personalized images based on the protected images as the generation performance is degraded by the cloak.
While these methods represent significant progress, they all follow an image-specific assumption that the cloaks generated by defenders are built upon a one-to-one correspondence with the protected images.
Given the vast amounts and rapid updates of personal images accessible online, the image-specific assumption may not be realistic in practice, since each new image necessitates the reapplication of these techniques to create new privacy cloaks, rendering them highly inefficient and burdensome for typical users.

In this paper, we first investigate the robustness of current defense approaches.
As shown in Fig.~\ref{fig:intro} (a), when applying cloaks generated by image-specific protection methods to other images of the same identity, the protective performance is notably weakened. 
This brings us to the question:
\textbf{\emph{How can we design a universal privacy cloak that can protect all images of an identity?}}
This demand poses two challenges: 
1) ``what to protect''. 
In the traditional ``image-specific'' setting, protection is applied to a predefined set of images, where the defender has full knowledge of the protection targets.
However, in the ``identity-specific'' setting, the exact distribution of images requiring protection is unknown. Defenders are only provided with scarce samples from the target identity, which is insufficient to capture the underlying image distribution that needs to be protected.
2) ``how to protect''. 
As our objective shifts from optimizing cloaks for specific ``images'' to broader ``identities'', 
how to optimize cloaks in an ``identity-specific'' manner remains an open question that requires further investigation.

To tackle these challenges, we propose ID-Cloak to generate identity-specific universal cloaks from only a few images of an individual. Specifically, we solve the ``what to protect'' problem by learning an identity subspace in the text embedding space. This identity subspace is intended to capture the entire distribution of the person, and implicitly covers all possible images to be protected, as illustrated in Fig.~\ref{fig:intro} (b).
Specifically, we model the subspace as a Gaussian distribution, where the mean and variance are estimated from a set of anchor points in the text embedding space. These anchor points are learned via prompt learning, which capture both core identity and diverse protection contexts.
To solve the ``how to protect'' question, we develop a novel optimization objective that encourages the cloak to steer the model away from its normal output within this modeled subspace. 

Qualitative and quantitative experiments demonstrate that our method can learn an identity-specific cloak from a small set of images, effectively protecting all possible images of the target identity. Our contributions are as follows:   
\begin{itemize}
    \item We introduce a novel protection paradigm against personalization misuse, shifting from image-specific to identity-specific defenses for practical usability.
    \item We propose ID-Cloak to craft identity-specific cloaks from a minimal set of an individual's images. It first models an identity subspace to capture the underlying protection distribution, then optimizes the cloak within this subspace using a novel optimization objective.
    \item Our experimental results, both qualitative and quantitative, demonstrate that our approach achieves robust protection across all possible images of an individual using a single universal mask.
\end{itemize}

\section{Related Work}
\subsection{Personalized T2I Generation}
With the advance of diffusion models, current text-to-image (T2I) generation~\cite{nichol2021glide, rombach2022high, wang2024stablegarment, cui2024localize} has shown remarkable generalization ability. 
As these methods ignore the concepts that do not appear in the training set, some works study personalized text-to-image generation which aims to adapt text-to-image models to specific concepts (attributions, styles, or objects) given several reference images.
For example, Textual Inversion~\cite{gal2022image} adjusts text embeddings of a new pseudoword to describe the concept. DreamBooth~\cite{ruiz2023dreambooth} fine-tunes denoising networks to connect the novel concept and a less commonly used word token. 
Based on that, several recent works~\cite{kumari2023multi,chen2023disenbooth, shi2024instantbooth} have been proposed to enhance controllability and flexibility when processing image visual concepts. 
These advancements enhance the capabilities of text-to-image models, making them more accessible to a wider range of users.

\subsection{Adversarial Examples}
Adversarial examples are crafted by adding imperceptible perturbations to mislead models, primarily applied in anti-classification, anti-deepfakes, and anti-facial recognition. Existing methods fall into two categories:  
\textbf{Image-specific adversarial examples} generate tailored perturbations per image. \citet{szegedy2013intriguing} pioneered this concept with LBFGS optimization, while \citet{goodfellow2014explaining} proposed the efficient FGSM. Subsequent works~\cite{xiao2018generating, xiong2023black, chen2023advdiffuser} improved naturalness via generative models. These are extended to disrupt deepfakes~\cite{ruiz2020disrupting, wang2022anti, wang2022deepfake, li2023unganable} and protect facial privacy~\cite{shan2020fawkes, yang2021towards, cherepanova2021lowkey, deb2020advfaces} from unauthorized face recognition systems.  
\textbf{Universal adversarial perturbations (UAPs)} apply a single perturbation to all images. \citet{moosavi2017universal} first revealed UAPs' existence, with \citet{liu2023enhancing} addressing gradient vanishing via aggregation and \citet{poursaeed2018generative} synthesizing UAPs via generative models. 
For privacy, \citet{zhong2022opom} proposed gradient-based OPOM for identity-specific protection, while \citet{liu2025advcloak} trained generators for natural adversarial cloaks. Our work aligns with UAPs but focuses on protecting all images of a target identity from unauthorized personalized generation.

\subsection{Anti-Personalization}
The remarkable generative capability of personalized T2I generation comes with safety concerns~\cite{carlini2023extracting, vyas2023provable}, particularly regarding the unauthorized exploitation of personal images. 
To mitigate these risks, recent studies have proposed the use of adversarial examples to counteract such safety issues. 
AdvDM \cite{liang2023adversarial} pioneered a theoretical framework for crafting adversarial examples against diffusion models. 
Anti-DreamBooth \cite{van2023anti} tackled anti-personalization with a bi-level protection objective and ASPL optimization, later refined by \citet{wang2024simac} via time-step selection. 
MetaCloak \cite{liu2024metacloak} enhanced cloak robustness against image transformations using ensemble learning and EoT, while \citet{xue2023toward} reduced computational costs via SDS loss. 
\citet{li2024pid} and \citet{wan2024prompt} addressed prompt discrepancies between protectors and attackers with encoder-based protection and prompt distribution modeling, respectively.
Despite these advancements, existing methods predominantly generate image-specific cloaks, which are impractical for widespread user adoption. 
In contrast, our work introduces a universal cloak tailored to individual users, enabling it to be applied across all their images, significantly enhancing usability and reducing privacy risks.

\section{Method}

\subsection{Problem Definition}
We consider a scenario where a user \( k \) (referred to as the protector) aims to safeguard all of their current and potential future images, which are modeled as samples from the distribution \( q(x) \). The user's objective is to prevent unauthorized attackers from utilizing any of their images to train personalized models for customized image generation.
To achieve this, the user seeks to create a personal universal cloak \( \delta \), which can be applied to any image \( x \sim q(x) \). Specifically, the user applies the cloak to their images, resulting in perturbed images \( x' = x + \delta \), which are then published online. The set of published images on the Internet is denoted as \( X_p = \{x_1', x_2', \ldots, x_i', \ldots\} \). Subsequently, an attacker may attempt to extract these images \( X_p \) to train personalized models.
The user's goal is to ensure that personalized models trained on the protected images exhibit degraded generation quality. Formally, this objective is defined as:
\begin{equation}
\begin{aligned}
\delta^{*} &= \underset{\delta}{\arg\min} \; \mathcal{A}(\theta^{*}, k) \\
\text{subject to} \quad & \theta^{*} = \underset{\theta}{\arg\min} \; \mathbb{E}_{x' \sim X_p} \left[ \mathcal{L}_{p}(\theta, x') \right], \\
\text{and} \quad & \| \delta \|_{p} \leq \eta.
\end{aligned}
\end{equation}
where \(\theta\) denotes a pre-trained text-to-image model, \( \mathcal{L}_{p} \) represents the personalized training objective, and \( \mathcal{A}(\theta^{*}, k) \) is an evaluation function that assesses the quality of images generated by the personalized model \( \theta^{*} \) with respect to the protected identity \( k \). To ensure visual imperceptibility, the cloak \(\delta\) is constrained within an \(\ell_p\)-ball of radius \(\eta\).

\subsection{ID-Cloak: Crafting Identity-Specific Cloaks}
In the above problem formulation, a critical aspect is the characterization of the 
personal image distribution \( q(x) \). Since the real distribution is unavailable, We can only describe this distribution based on the available set of personal face images \(X_c \sim q(x)\) provided by the user. While a larger set of images would allow for a more accurate estimation of the distribution, obtaining a vast number of individual images is often impractical and contrary to our initial objective. Therefore, we aim to estimate \( q(x) \) using \(X_c=\{x_i\}^N_{i=1}\) with a limited number of images \(N\).

The fundamental intuition is that a more precise approximation of the personal image distribution enhances the universality of the cloak, thereby improving its transferability across different images.
Building on this, we propose ID-Cloak, a novel method for generating such identity-specific universal cloaks using a small set of an individual's images. Our approach comprises two main stages:
1) identity subspace modeling: utilizing the input few-shot images, we learn a subspace in the text embedding space which represents the individual. This subspace, together with a T2I generative model, is intended to capture the entire personal image distribution for protection.
2) universal cloak optimization: based on the modeled subspace, we develop an optimization objective to that encourages the cloak to steer the model away from its normal output within
the subspace.
\subsubsection{Modeling the Identity Subspace}

We base our approach on the following assumption: Let an individual \( k \) possess a protected image distribution \( q(x) \) defined over the image space \( \mathcal{X} \). Consider a text-to-image diffusion model with parameters \( \theta \), characterized by its conditional sampling distribution \( p_\theta(x|c) \), where \( c \in \mathcal{C} \) denotes text conditions in the text embedding space \( \mathcal{C} \). We hypothesize the existence of a latent identity subspace \( Q(c) \) defined over \( \mathcal{C} \) that encapsulates all text conditions semantically associated with the protected identity of individual \( k \). By combining \(  Q(c) \) with the conditional sampling distribution \( p_\theta(x|c) \), we can approximate the image distribution \( q(x) \) as:
\begin{equation}
    p_\theta(x) = \int p_\theta(x | c) Q(c) \, dc.
\end{equation}
This formulation reframes the problem by shifting the focus from complex image distributions to a more structured and interpretable text-based representation. Specifically, it allows us to model the protected identity's image distribution by focusing on a semantically meaningful subspace in the text embedding space.

The ideal subspace \( Q(c) \) should capture both the commonalities (core identity information of the individual) and the variations (diversity of protection contexts, such as backgrounds, poses, illuminations, and expressions) to cover all potential protection scenarios. 
To approximate \( Q(c) \), we model it as a Gaussian distribution \( \hat{Q}(c) \), parameterized by the mean and variance of a set of anchor points in the text embedding space. These anchor points are learned via prompt tuning, initialized from the core identity point and optimized to associate with specific image instances, thereby incorporating diverse protection contexts. 
We adopt this approximation for two reasons:
1) Gaussian distribution effectively models large sample sizes, ideal for subspace representation.
2) It aligns with our ideal text embedding distribution, concentrated around the core identity point.

\begin{algorithm}[t]
    \caption{Learning Identity Subspace}
    \label{alg:1}
\begin{algorithmic}[1] 
\REQUIRE Personal images \(X_c=\{x_i\}^N_{i=1} \), diffusion model \( \theta \) with text encoder \(\tau\), identity learning steps \( C \), prompt tuning steps \( M \), identity descriptor \( V^* \)

\STATE // Step 1: identity token learning
\STATE Construct textual description \(P\) using \(V^*\)
\FOR{$i = 1$ {\bfseries to} $C$}
    \STATE Sample \( x \sim X_c, \epsilon \sim \mathcal{N}(\mathbf{0,I}),t \in U(1, T) \)
    \STATE \(x_t=\sqrt{\alpha_t} x_0 + \sqrt{1-\alpha_t}\epsilon\)
    \STATE Take a gradient step on\( \nabla_{\theta,\tau}\| \epsilon - \epsilon_\theta(x_t, t, \tau(P)) \|_2^2 \) 
\ENDFOR
\STATE {\bfseries Yield:} personalized model \( \theta^* \), text encoder \( \tau^* \)

\STATE // Step 2: context diversification
\STATE Obtain core identity point \( c_{ID} = \tau^*(P) \)
\STATE Initialize anchor points \( \{c_i =  c_{ID} \}_{i=1}^{N}\)
\FOR{$i = 1$ {\bfseries to} $M$}
    \STATE Sample \(\epsilon \sim \mathcal{N}(\mathbf{0,I}),t \in U(1, T) \)
    \FOR{$j = 1$ {\bfseries to} $N$}
        \STATE \( c_{j} \gets c_{j} - \nabla_{c_j} \| \epsilon - \epsilon_{\theta^*}\left(x_{j,t}, t, c_j\right) \|_2^2 \)
    \ENDFOR
\ENDFOR
\STATE Construct identity subspace 

\( \hat{Q}(c) \sim \mathcal{N}(\mu(  \{c_i\}^N_{i=1}), \sigma( \{c_i\}^N_{i=1}) )\) 
\STATE {\bfseries Return:} Identity subspace \( \hat{Q} \), personalized model \( \theta^* \)
\end{algorithmic}
\end{algorithm}

We begin by learning the core identity information of the individual from a few-shot set of input images \(X_c=\{x_i\}^N_{i=1}\).
To achieve this, we define a unique identifier \( V^*\) that represents the person's identity. This identifier is combined with a sentence template to form a textual description \(P\) (e.g., ``a photo of \(V^*\) person'').
We implant the identity information into \( V^*\) by optimizing the following objective, as described in~\cite{ruiz2023dreambooth}:
\begin{equation}
\min_{\theta,\tau} \mathbb{E}_{x \sim {X_c}, \epsilon , t } 
\left[ \| \epsilon - \epsilon_\theta(x_t, t, \tau(P)) \|_2^2 \right],
\end{equation}
where $\tau(\cdot)$ is a text encoder producing text embeddings.
In this stage, we optimize the full model parameters \(\theta\) with the text encoder \(\tau\) to ensure high fidelity to the original identity, resulting in a personalized model \(\theta^*\)  with text encoder \(\tau^*\).

Subsequently, to exploit the diversity inherent in the input image set, we optimize a set of soft embeddings \( \{c_i\}_{i=1}^N \), where each \( c_i \) is associated with a specific image \( x_i \) in the input set. These embeddings are initialized with \( c_{ID} \), which represents the core identity learned from the previous stage: \(c_{ID} = \tau^*(P) \).
Our goal is to compute \( \{c_i\}_{i=1}^N \) such that each \( c_i \) best describes its corresponding image \( x_i \). Formally, this can be expressed as:
\begin{equation}
c_i = \arg\max_c p(c | x_i, c_{ID}).
    \label{eq:context_diverse_vanilla}
\end{equation}
However, directly maximizing this likelihood is intractable. Following prior work \cite{chen2023robust,wan2024prompt,zhang2025generate}, we reformulate the problem as an expectation minimization task (refer Appendix~\ref{sec:derivation} for details):
\begin{equation}
    \min_{\{c_i\}_{i=1}^N} \mathbb{E}_{\epsilon,t } \sum_{i=1}^N \left\| \epsilon - \epsilon_{\theta^*}\left(x_{i,t}, t, c_i\right) \right\|_2^2.
    \label{eq:context_diverse_after}
\end{equation}
Finally, using the obtained soft embeddings \( \{c_i\}_{i=1}^N \), we approximate the identity subspace as a Gaussian distribution:
\begin{equation}
\hat{Q}(c) \sim \mathcal{N}\left(c; \mu(\{c_i\}_{i=1}^N), \sigma(\{c_i\}_{i=1}^N)\right),
\end{equation}
where \( \mu(\{c_i\}_{i=1}^N) \) and \( \sigma(\{c_i\}_{i=1}^N) \) are the mean and standard deviation estimated in the text encoder space.
Once the subspace \( Q \) is established, we approximate the target distribution \( q(x) \) by sampling images from the subspace using the diffusion model. This is achieved through a Monte Carlo sampling approach: \(x \sim p_{\theta^*}(x | c)\), where \(c \sim Q(c)\).

\subsubsection{Optimizing Identity-Specific Cloaks}
In the previous section, we approximated the real personal image distribution $q(x)$ using a personalized diffusion model $\theta^*$ parameterized by an identity subspace $Q(c)$. Our current objective is to optimize a universal cloak $\delta$ to ensure that the cloak remains adversarial across the entire personal image distribution $q(x)$.
Formally, we seek for a cloak $\delta$ that maximizes the divergence between the model's output distribution under the cloaked input $p_{\theta^{*}}(x+\delta)$ and the personal image distribution $q(x)$. This is achieved by maximizing the cross-entropy between the two distributions:
\begin{equation}
    \delta := \arg\max_\delta -\mathbb{E}_{q(x)} \log p_{\theta^{*}}(x+\delta).
\end{equation}
This objective is closely aligned with the standard training objective of diffusion models \cite{sohl2015deep,ho2020denoising}, but with an adversarial intent: instead of learning to generate samples from $q(x)$, we aim to disrupt the model's ability to accurately represent the personal image distribution.
Following \citet{ho2020denoising}, we reformulate the cross-entropy objective into a tractable denoising score matching loss:
\begin{equation}
\begin{aligned}
\max_{\delta} \; & \mathbb{E}_{x\sim q(x), \epsilon, t} \left[ \| \epsilon - \epsilon_{\theta^*}(x_t', t, c) \|_2^2 \right],  \\
\end{aligned}
\end{equation}
where $x_t' = \sqrt{\alpha_t}(x+\delta) + \sqrt{1-\alpha_t}\epsilon$.
However, directly optimizing this objective requires sampling from \(q(x)\) , which is intractable. Instead, we leverage the fact that the personalized model \(\theta^*\)—trained to approximate \(q(x)\) via its subspace \(Q(c)\)—provides an accessible surrogate distribution \(p_{\theta^*}(x|c)\). By sampling from \(p_{\theta^*}(x|c)\) where \(c \sim Q(c)\), we can effectively approximate \(q(x)\). This transforms the objective into a tractable form:
\begin{equation}
\resizebox{.9\hsize}{!}{$\displaystyle
\begin{aligned}
    \max_{\delta} \; & \mathbb{E}_{x\sim p_{\theta^*}(x|c), c\sim Q(c), \epsilon, t} \left\| \epsilon_{\theta^*}(x_t, t, c) - \epsilon_{\theta^*}(x'_t, t, c) \right\|_2^2,
\end{aligned}
$}
\label{eq:vanilla_obj}
\end{equation}
where $x_t = \sqrt{\alpha_t} x + \sqrt{1-\alpha_t} \epsilon$.
Directly optimization of Eq.~(\ref{eq:vanilla_obj}) requires sampling from the full reverse diffusion chain \(   p(x_T) \prod_{t=1}^{T} p_\theta(x_{t-1} | x_t) \), which incurs prohibitive computational costs.
Observing that while the cloak $\delta$ needs to be ultimately applied to the clean image $x$, the other terms in Eq.~(\ref{eq:vanilla_obj}) only depend on the intermediate noisy latent $x_t$ sampled from the reverse process $T\rightarrow t$. 
To bypass the costly reverse process (\(t \rightarrow 0\)) and redundant forward passes (\(0 \rightarrow t\)), we propose a one-step latent cloaking strategy leveraging DDIM~\cite{song2020denoising}.
Specifically, given a noisy latent $x_t$, we first estimate a clean image $\hat{x}_0$ through deterministic denoising:
\begin{equation}
\hat{x}_0=\frac{x_t-(\sqrt{1-\alpha_t})\epsilon_{\theta^*}({x}_t,t,c)}{\sqrt{\alpha_t}}.
\end{equation}
Next, the cloak $\delta$ is applied to $\hat{x}_0$:
\begin{equation}
\hat{x}_0' = \hat{x}_0 + \delta .
\end{equation}
The perturbed estimate $\hat{x}_0'$ is then reprojected to the noisy latent space at timestep $t$ through:
\begin{equation}
\hat{x}_t' = \sqrt{\alpha_t} \hat{x}_0' + \sqrt{1-\alpha_t} \epsilon_{\theta^*}({x}_t,t,c).
\end{equation}
Finally, our target becomes:
\begin{equation}
    \max_{\delta} \mathbb{E}_{x_t, c\sim Q(c), t} \left[\left\| \epsilon_\theta(x_t, t, c) - \epsilon_\theta(\hat{x}_t', t, c) \right\|_2^2\right].
    \label{eq:final_obj}
\end{equation}

By maximizing the discrepancy between noise predictions for the original and perturbed latents, the cloak is encouraged to guide the model output away from its normal behavior within the subspace across diffusion timesteps, ultimately causing the final generated images to deviate significantly from the original ones. 
The algorithm is outlined in Alg.~(\ref{alg:2}).
When updating the cloak, the Projected Gradient Descent (PGD) \cite{madry2017towards} and Stochastic Gradient Aggregation (SGA) technique \cite{liu2023enhancing} are employed to improve gradient stability and enhance the optimization efficacy.
Further details are provided in the Appendix~\ref{sec:gradient_update}.

\begin{algorithm}[t]
    \caption{Optimizing Identity-specific Universal Cloak}
    \label{alg:2}
\begin{algorithmic}[1] 
\REQUIRE Customized diffusion model \( \theta^* \), personal subspace \( \hat{Q}(c) \), training iterations \( N \), noise budget \( \eta \), PGD step size \( \alpha \)

\STATE {\bfseries Initialize:} \( \delta= 0 \)
\FOR{$n = 1$ {\bfseries to} $N$} 
        \STATE Sample \( c \sim \hat{Q}(c), t \in U(0, T) \)
        \STATE Sample \( x_t = \texttt{sample}(\theta^*, t, c) \)
        \STATE Obtain \( \hat{x}_t' = \texttt{applyCloak}(x_t, \delta) \)
        \STATE Compute grad \(g = \nabla_{\hat{x}_t'}\| \epsilon_{\theta^*}(x_t, t, c) - \epsilon_{\theta^*}(\hat{x}_t', t, c) \|_2^2\) 
    \STATE \( \delta \gets \text{clip}^\eta_\delta(\delta + \alpha \cdot \text{sgn}(g)) \) 
\ENDFOR
\STATE {\bfseries Return:} Universal cloak \( \delta^* \) for identity \( k \)
\end{algorithmic}
\end{algorithm}

\begin{figure*}[ht]
    \centering
    \includegraphics[width=1\linewidth]{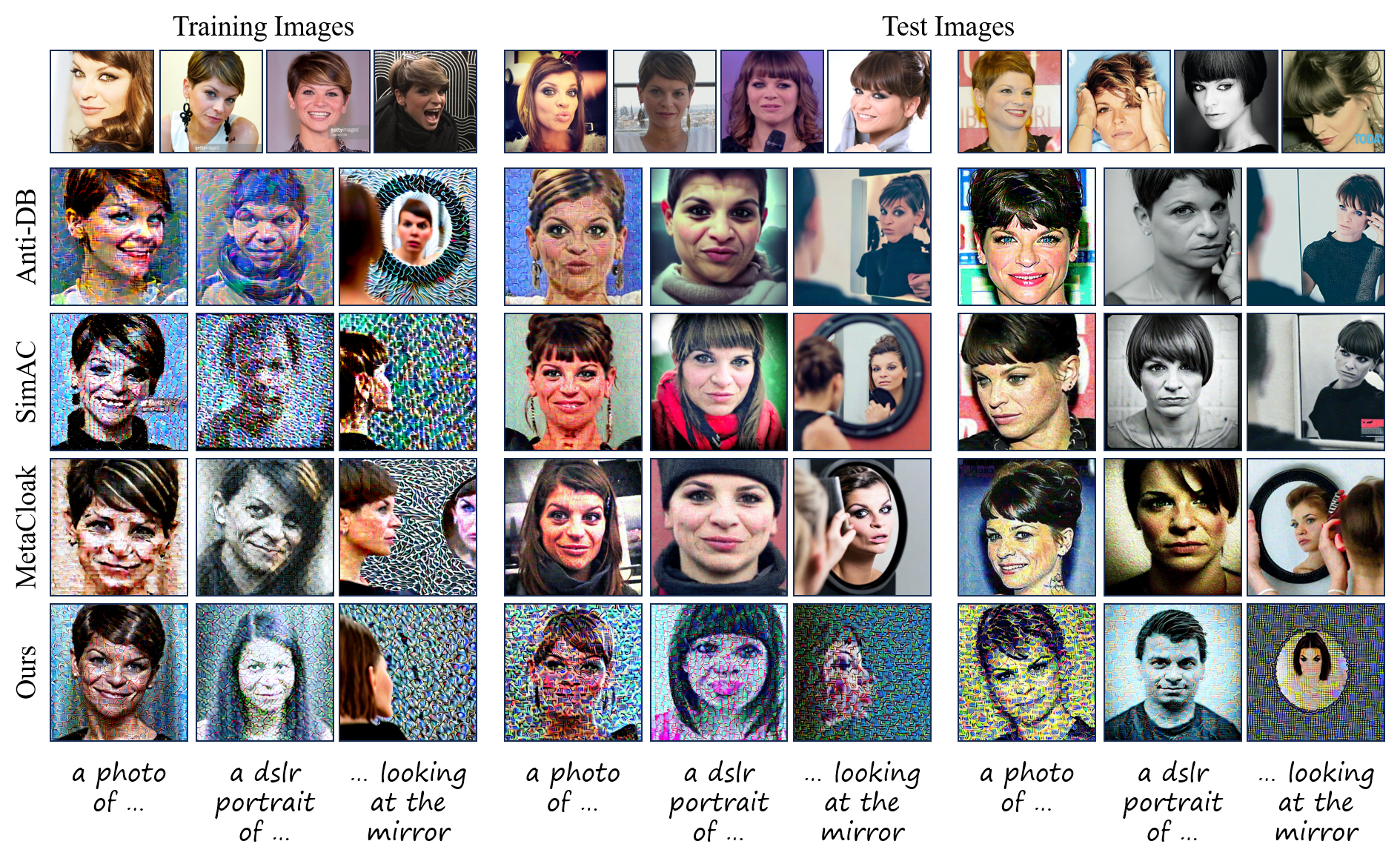}
    \vspace{-7mm}
    \caption{Qualitative results on VGGFace2 dataset. The cloaks are generated from the images of training set, then applied on the same training set and different test sets respectively. Each row represents a method, and each column represents a different test prompt.}
    \label{fig:main_result}
\end{figure*}

\section{Experiment}
\begin{table*}
\centering
\caption{Comparison with other open-sourced anti-personalization methods on VGGFace2 (left) and CelebA-HQ (right). We evaluate the performance under three different prompts during personalization. \textit{Universal} versions are denoted with ``+''.}
\setlength{\tabcolsep}{10pt}
\resizebox{0.95\textwidth}{!}{%
\begin{tabular}{lcccc|cccc}
\toprule
\multicolumn{1}{c}{\multirow{2}{*}{\textbf{Method}}} & \multicolumn{4}{c}{\textbf{VGGFace2}} & \multicolumn{4}{c}{\textbf{CelebA-HQ}} \\
\cmidrule(lr){2-5} \cmidrule(lr){6-9}
 & \textbf{BRISQUE$\uparrow$} & \textbf{ISM$\downarrow$} & \textbf{FDFR$\uparrow$} & \textbf{SER-FIQ$\downarrow$} &
\textbf{BRISQUE$\uparrow$} & \textbf{ISM$\downarrow$} & \textbf{FDFR$\uparrow$} & \textbf{SER-FIQ$\downarrow$} \\

\midrule

&\multicolumn{4}{c@{\quad}}{\textit{``a photo of sks person''}} & \multicolumn{4}{c}{\textit{``a photo of sks person''}} \\
Anti-DB        & 32.132 & 0.600 & 0.004 & 0.730 & 29.160 & 0.543 & 0.002 & 0.707 \\
SimAC          & 34.937 & 0.590 & 0.007 & 0.731 & 28.664 & 0.519 & 0.003 & 0.698 \\
MetaCloak      & 26.616 & 0.524 & 0.008 & 0.699 & 30.074 & 0.461 & 0.007 & 0.658 \\
Anti-DB+       & 35.493 & 0.577 & 0.008 & 0.722 & 31.280 & 0.447 & 0.006 & 0.648 \\
SimAC+         & 38.164 & 0.575 & 0.010 & 0.694 & 30.409 & 0.442 & 0.007 & 0.644 \\
MetaCloak+     & 24.550 & 0.507 & 0.009 & 0.674 & 28.719 & \textbf{0.416} & 0.007 & 0.637 \\
Ours           & \textbf{38.472} & \textbf{0.469} & \textbf{0.143} & \textbf{0.557} & \textbf{38.599} & 0.477 & \textbf{0.025} & \textbf{0.632} \\
\midrule

&\multicolumn{4}{c@{\quad}}{\textit{``a dslr portrait of sks person''}} & \multicolumn{4}{c}{\textit{``a dslr portrait of sks person''}} \\
Anti-DB        & 10.519 & 0.434 & 0.012 & 0.714 & 5.008 & 0.432 & 0.004 & 0.749 \\
SimAC          & 10.688 & 0.444 & 0.021 & 0.708 & 4.751 & 0.432 & 0.007 & 0.747 \\
MetaCloak      & 12.461 & 0.447 & 0.013 & 0.701 & 10.237 & 0.424 & 0.011 & 0.750 \\
Anti-DB+       & 11.807 & 0.414 & 0.016 & 0.706 & 6.863 & 0.427 & 0.004 & 0.747 \\
SimAC+         & 18.466 & 0.424 & 0.043 & 0.690 & 6.368 & 0.446 & 0.005 & 0.754 \\
MetaCloak+     & 11.443 & 0.449 & 0.013 & 0.706 & 9.501 & 0.412 & 0.018 & 0.746 \\
Ours           & \textbf{26.143} & \textbf{0.336} & \textbf{0.228} & \textbf{0.557} & \textbf{21.304} & \textbf{0.363} & \textbf{0.055} & \textbf{0.684} \\
\midrule

&\multicolumn{4}{c@{\quad}}{\textit{``a photo of sks person looking at the mirror''}} & \multicolumn{4}{c}{\textit{``a photo of sks person looking at the mirror''}} \\
Anti-DB        & 15.427 & 0.400 & 0.064 & 0.549 & 15.189 & 0.420 & 0.055 & 0.586 \\
SimAC          & 22.163 & 0.425 & 0.048 & 0.563 & 17.559 & 0.413 & 0.055 & 0.589 \\
MetaCloak      & 20.221 & 0.413 & 0.060 & 0.564 & 22.009 & 0.429 & 0.053 & 0.585 \\
Anti-DB+       & 18.423 & 0.405 & 0.069 & 0.543 & 17.577 & 0.406 & 0.051 & 0.584 \\
SimAC+         & 27.421 & 0.388 & 0.073 & 0.531 & 20.483 & 0.402 & 0.053 & 0.568 \\
MetaCloak+     & 19.526 & 0.413 & 0.067 & 0.557 & 21.920 & 0.418 & 0.057 & 0.581 \\
Ours           & \textbf{28.537} & \textbf{0.288} & \textbf{0.259} & \textbf{0.388} & \textbf{28.375} & \textbf{0.340} & \textbf{0.111} & \textbf{0.498} \\
\bottomrule
\end{tabular}%
}
\label{tab:merged_results}
\end{table*}

\subsection{Experiment Setup}
\textbf{Datasets.} 
We select two face datasets: CelebA-HQ~\cite{liu2015deep} and VGGFace2~\cite{cao2018vggface2}. 
For a comprehensive evaluation, we randomly select 50 identities from each dataset. For each identity, we randomly pick 12 images and split them into two subsets: a training set and a test set. The training set is used to generate the adversarial privacy cloak, while the test set is used to evaluate the protection performance of the generated cloak. The training set and test set comprise 4 images and 8 image respectively. 

\textbf{Evaluation metrics.}
Our method aims to disrupt target personalized models, causing them to generate poor-quality images of the protected identity. To evaluate the effectiveness, we design metrics that assess two key aspects of distortion: semantic-related distortion and quality-related distortion.
For semantic-related distortion, our goal is to significantly alter the subject's identity in generated images to prevent misuse. We first evaluate subject detectability using the Face Detection Failure Rate (FDFR) with RetinaFace~\cite{deng2020retinaface} as the detector. If a face is detected, we then measure semantic similarity via Identity Score Matching (ISM), which computes cosine similarity between ArcFace embeddings of the generated and original faces.
For quality-related distortions, we aim to degrade generated image quality.
To measure this, we adopt two metrics: BRISQUE~\cite{mittal2012no}, widely used for image quality assessment, and SER-FIQ~\cite{terhorst2020ser}, designed for facial image quality evaluation.
To measure the defense’s effectiveness, we generate 30 images using three different prompts for each trained personalized model.

\textbf{Implementation details.}
For a fair comparison, we set the same noise budget for all methods, with \(\eta=16/255\), and the optimization steps and step sizes are aligned with the optimal settings specified in each baseline. The default base model used across all methods is Stable Diffusion v2.1.
For our method, in the stage of learning the personal subspace, we first fine-tune both the U-Net and text encoder for 1000 steps to learn identity information, using a learning rate of $1e-5$. 
Subsequently, we perform prompt tuning for 50 steps with a learning rate of $1e-3$. 
In the stage of optimizing universal privacy cloaks, we employ DDIM for the sampling process in the diffusion model, with a total of 50 sampling steps. For updating the universal cloak, we set the total training iterations to 200, with 10 inner iterations for gradient aggregation in each iteration, and the step size $\alpha$ is set to 0.05. 
We use the protected images with cloaks for personalized text-to-image generation, adopting DreamBooth as the default personalization technique. After fine-tuning for 1,000 steps, we generate images to measure the defense performance.

\textbf{Baselines.}
We conduct a comprehensive comparison of our proposed method against several state-of-the-art baselines, including Anti-DreamBooth~\cite{van2023anti}, SimAC~\cite{wang2024simac}, and MetaCloak~\cite{liu2024metacloak}. Notably, these existing methods primarily concentrate on generating image-specific privacy cloaks. To make a fair comparison, we extend these methods to generate universal cloaks by employing a gradient-averaging update strategy~\cite{moosavi2016deepfool, moosavi2017universal}. We denote these improved variants as \textit{Universal} methods and their original counterparts as \textit{Image-specific} methods.
For the \textit{Image-specific} methods, we randomly transfer the cloaks learned from training images to other images in the test set, thereby constructing the final protected images for personalized fine-tuning. In contrast, for the \textit{Universal} methods, we directly apply the cloak generated on the training set to the images in the test set, yielding the final protected images for personalized fine-tuning.

\subsection{Comparison with Baseline Methods}
Qualitative results are presented in Figure~\ref{fig:main_result}. 
We conduct experiments on both the training and test sets. 
It is evident that cloaks generated by other methods are effective only on the training set which is the default setting in previous methods.
When applied to other images of the same individual, the protective effectiveness diminishes significantly or is even completely lost. In contrast, our method consistently provides effective protection across both the training and test sets. 
The results with different prompts further demonstrate the robustness of our method.

The quantitative comparisons on the VGGFace2 dataset and the CelebA-HQ dataset are presented in Table~\ref{tab:merged_results}. Our method consistently outperforms all baseline approaches across all prompts and datasets. Notably, the \textit{universal} cloak generation variants exhibit superior performance compared to their original \textit{image-specific} counterparts. This finding suggests that, under the setting of learning identity-specific universal cloaks, learning a single universal cloak is more effective than learning image-specific cloaks. However, our method still demonstrates a significant performance improvement over these \textit{universal} methods.
For instance, in the critical metric of FDFR, which measures whether a face can be detected in the generated image, our method, ID-Cloak, achieves an average improvement factor of \textbf{5.0} and \textbf{2.4} compared to the previous state-of-the-art (SoTA) methods on VGGFace2 and CelebA-HQ datasets, respectively. Even when a face is detectable, the ISM and SER-FIQ metrics indicate that our ID-Cloak generates faces with the greatest identity deviation from the original, while achieving the lowest quality for the face portion of the image. Additionally, the results on the BRISQUE metric suggest that ID-Cloak effectively degrades the overall image quality of the generated images.
These results validate our method's effectiveness in creating identity-specific cloaks for robust facial privacy protection and demonstrate strong generalization across individual face images.

\subsection{Transfer Experiments}
In real-world applications, attackers may employ models or personalization techniques different from those used by protectors.
In this section, we conduct a series of transferability experiments.
1) model transferability: we investigate whether the privacy cloaks generated for one model can effectively protect against exploitation by other models.
2) personalization techniques transferability: we analyze whether the cloaks remain effective when downstream attackers apply different personalization techniques.

\textbf{Model transferability.} 
To evaluate the transferability of the cloaks across different target models, we specifically investigate the transferability between Stable Diffusion v1.5 and Stable Diffusion v2.1. We assess the effectiveness of cloaks learned on Stable Diffusion v2.1 when applied to the personalization process on Stable Diffusion v1.5, and vice versa. The results are summarized in Table~\ref{tab:transfer_model}.
Our findings demonstrate that ID-Cloak exhibits strong transferability between these two model versions. Across all metrics, performance remains stable in different cross-model settings, with only slight degradation. This robustness can likely be attributed to the similarity in the condensed representations of images across the models. 
These results demonstrate the cloaks generated by our method provide robust, broad protection across different models.
\begin{table}
\centering
\caption{Transferability results across different models. We evaluate on Stable Diffusion 1.5 and Stable Diffusion 2.1 under three different prompts. }
\setlength{\tabcolsep}{10pt}
\resizebox{\linewidth}{!}{%
\begin{tabular}{cccccc}
\toprule
\multirow{2}{*}{Train} & \multirow{2}{*}{Test} & \multicolumn{4}{c}{\textit{``a photo of sks person''}} \\
\cmidrule(lr){3-6}
     &      & BRISQUE$\uparrow$ & ISM$\downarrow$ & FDFR$\uparrow$ & SER-FIQ$\downarrow$ \\
\midrule
v2.1 & v2.1 & 38.472 & 0.469 & 0.143 & 0.557 
\\
v2.1 & v1.5& 28.761 & 0.389 & 0.426 & 0.411 
\\
v1.5& v2.1 & 37.231 & 0.470 & 0.124 & 0.546 
\\
\midrule
\multirow{2}{*}{Train} & \multirow{2}{*}{Test} & \multicolumn{4}{c}{\textit{``a dslr portrait of sks person''}} \\
\cmidrule(lr){3-6}
     &      & BRISQUE$\uparrow$ & ISM$\downarrow$ & FDFR$\uparrow$ & SER-FIQ$\downarrow$ \\
\midrule
v2.1 & v2.1 & 26.143 & 0.336 & 0.228 & 0.557 
\\
v2.1 & v1.5& 11.026 & 0.375 & 0.012 & 0.634 
\\
v1.5& v2.1 & 24.139 & 0.343 & 0.176 & 0.586 
\\
\midrule
\multirow{2}{*}{Train} & \multirow{2}{*}{Test} & \multicolumn{4}{c}{\textit{``a photo of sks person looking at the mirror''}} \\
\cmidrule(lr){3-6}
     &      & BRISQUE$\uparrow$ & ISM$\downarrow$ & FDFR$\uparrow$ & SER-FIQ$\downarrow$ \\
\midrule
v2.1 & v2.1 & 28.537 & 0.288 & 0.259 & 0.388 
\\
v2.1 & v1.5& 26.695 & 0.343 & 0.193 & 0.488 
\\
v1.5& v2.1 & 27.326 & 0.300 & 0.240 & 0.406 
\\
\bottomrule
\end{tabular}%
}
  \label{tab:transfer_model}
\end{table}

\textbf{Personalization techniques transferability.}
To evaluate the robustness of the proposed method against different personalization techniques, we apply ID-Cloak to three widely adopted personalization techniques: Dreambooth~\cite{ruiz2023dreambooth}, Dreambooth with LoRA~\cite{hu2021lora} and Textual Inversion~\cite{gal2022image}.
Dreambooth corresponds to the default configuration of our above experiments. LoRA is a widely adopted low-rank personalization method suited for low computational resources. Textual Inversion customizes concepts by optimizing a word vector rather than fine-tuning the entire model.
As shown in Table~\ref{tab:transfer_ft_technique}, our method, ID-Cloak, effectively defends against both Dreambooth, LoRA and Textual Inversion, highlighting its efficacy in countering various personalization techniques.

\begin{table}[ht]
\centering
\caption{Transferability results across different personalization techniques. We provide the results under three different prompts.}
\setlength{\tabcolsep}{10pt}
\resizebox{\linewidth}{!}{%
\begin{tabular}{ccccc}
\toprule
\multirow{2}{*}{Method} & \multicolumn{4}{c}{\textit{``a photo of sks person''}} \\
\cmidrule(lr){2-5}
              & BRISQUE$\uparrow$ & ISM$\downarrow$ & FDFR$\uparrow$ & SER-FIQ$\downarrow$ \\
\midrule
DreamBooth& 38.472 & 0.469 & 0.143 & 0.557 
\\
DreamBooth-LoRA& 45.990 & 0.230 & 0.487 & 0.260 
\\
Textual Inversion& 59.309 & 0.273 & 0.411 & 0.560 
\\
\midrule
\multirow{2}{*}{Method} & \multicolumn{4}{c}{\textit{``a dslr portrait of sks person''}} \\
\cmidrule(lr){2-5}
              & BRISQUE$\uparrow$ & ISM$\downarrow$ & FDFR$\uparrow$ & SER-FIQ$\downarrow$ \\
\midrule
DreamBooth
& 26.143 & 0.336 & 0.228 & 0.557 
\\
DreamBooth-LoRA
& 31.812 & 0.189 & 0.277 & 0.452 
\\
Textual Inversion& 29.947 & 0.218 & 0.149 & 0.606 
\\
\midrule
\multirow{2}{*}{Method} & \multicolumn{4}{c}{\textit{``a photo of sks person looking at the mirror''}} \\
\cmidrule(lr){2-5}
              & BRISQUE$\uparrow$ & ISM$\downarrow$ & FDFR$\uparrow$ & SER-FIQ$\downarrow$ \\
\midrule
DreamBooth
& 28.537 & 0.288 & 0.259 & 0.388 
\\
DreamBooth-LoRA
& 31.129 & 0.139 & 0.359 & 0.265 
\\
Textual Inversion& 50.114 & 0.229 & 0.458 & 0.456 
\\
\bottomrule
\end{tabular}%
}
  \label{tab:transfer_ft_technique}
\end{table}

\subsection{Ablation Study}
\textbf{Effectiveness of proposed components.} 
To evaluate the individual contributions of ID-Cloak's components to its overall effectiveness, we conducted ablation studies on the VGGFace2 dataset.
We first ablated all components and directly optimized a single cloak using the input images via the gradient-averaging method~\cite{moosavi2016deepfool, moosavi2017universal}. Next, we tested a simplified version where a single point was used to represent the individual in the text embedding space, rather than modeling a subspace. 
The results, summarized in Table~\ref{tab:ablation}, demonstrate that all simpler or alternative configurations yield inferior performance compared to our full model. 
Specifically, the results of the second ablation study indicate that using a single point to describe an individual's identity lacks diversity and is prone to overfitting, failing to capture the full distribution of an individual's characteristics. In contrast, modeling a subspace by incorporating the diversity of protection contexts enables coverage of a broader range of potential protection scenarios, thereby enhancing the generalizability of the cloak.

\vspace{-2mm}
\begin{table}[h]
\centering
\caption{Ablation results of our proposed method.}
\setlength{\tabcolsep}{10pt}
\resizebox{\linewidth}{!}{%
\begin{tabular}{cc|cccc}
\toprule
Sub. & Obj. & BRISQUE$\uparrow$ & ISM$\downarrow$ & FDFR$\uparrow$ & SER-FIQ$\downarrow$ \\
\midrule
 &  & 22.71 & 0.468 & 0.035 & 0.649 \\
 & \checkmark  & 30.29 & 0.364 & 0.186 & 0.506 \\
\checkmark & \checkmark  & \textbf{31.05} & \textbf{0.364} & \textbf{0.210} & \textbf{0.501} \\
\bottomrule
\end{tabular}%
}
  \label{tab:ablation}
\end{table}

\section{Conclusion}
This paper introduces identity-specific cloaks (ID-cloaks), a novel privacy protection paradigm against misuse in personalized text-to-image generation. We formalize the task and propose an effective method for generating such cloaks. It first models an identity subspace in the text conditioning space to approximate the protection distribution, then optimizes universal masks utilizing a novel objective. Extensive experiments demonstrate the effectiveness of our solution, offering a scalable and practical advancement in privacy protection for generative models.

\bibliography{example_paper}

\begin{thebibliography}{50}
\providecommand{\natexlab}[1]{#1}
\providecommand{\url}[1]{\texttt{#1}}
\expandafter\ifx\csname urlstyle\endcsname\relax
  \providecommand{\doi}[1]{doi: #1}\else
  \providecommand{\doi}{doi: \begingroup \urlstyle{rm}\Url}\fi

\bibitem[Cao et~al.(2018)Cao, Shen, Xie, Parkhi, and Zisserman]{cao2018vggface2}
Cao, Q., Shen, L., Xie, W., Parkhi, O.~M., and Zisserman, A.
\newblock Vggface2: A dataset for recognising faces across pose and age.
\newblock In \emph{IEEE FG}, 2018.

\bibitem[Carlini et~al.(2023)Carlini, Hayes, Nasr, Jagielski, Sehwag, Tramer, Balle, Ippolito, and Wallace]{carlini2023extracting}
Carlini, N., Hayes, J., Nasr, M., Jagielski, M., Sehwag, V., Tramer, F., Balle, B., Ippolito, D., and Wallace, E.
\newblock Extracting training data from diffusion models.
\newblock In \emph{USENIX Security}, 2023.

\bibitem[Chen et~al.(2023{\natexlab{a}})Chen, Zhang, Wu, Wang, Duan, Zhou, and Zhu]{chen2023disenbooth}
Chen, H., Zhang, Y., Wu, S., Wang, X., Duan, X., Zhou, Y., and Zhu, W.
\newblock Disenbooth: Identity-preserving disentangled tuning for subject-driven text-to-image generation.
\newblock In \emph{ICLR}, 2023{\natexlab{a}}.

\bibitem[Chen et~al.(2024)Chen, Dong, Wang, Yang, Duan, Su, and Zhu]{chen2023robust}
Chen, H., Dong, Y., Wang, Z., Yang, X., Duan, C., Su, H., and Zhu, J.
\newblock Robust classification via a single diffusion model.
\newblock In \emph{ICML}, 2024.

\bibitem[Chen et~al.(2023{\natexlab{b}})Chen, Gao, Zhao, Ye, and Xu]{chen2023advdiffuser}
Chen, X., Gao, X., Zhao, J., Ye, K., and Xu, C.-Z.
\newblock Advdiffuser: Natural adversarial example synthesis with diffusion models.
\newblock In \emph{ICCV}, 2023{\natexlab{b}}.

\bibitem[Cherepanova et~al.(2021)Cherepanova, Goldblum, Foley, Duan, Dickerson, Taylor, and Goldstein]{cherepanova2021lowkey}
Cherepanova, V., Goldblum, M., Foley, H., Duan, S., Dickerson, J., Taylor, G., and Goldstein, T.
\newblock Lowkey: Leveraging adversarial attacks to protect social media users from facial recognition.
\newblock In \emph{ICLR}, 2021.

\bibitem[Cui et~al.(2024)Cui, Li, Li, Liu, Zou, and He]{cui2024localize}
Cui, X., Li, P., Li, Z., Liu, X., Zou, Y., and He, Z.
\newblock Localize, understand, collaborate: Semantic-aware dragging via intention reasoner.
\newblock In \emph{NeurIPS}, 2024.

\bibitem[Deb et~al.(2020)Deb, Zhang, and Jain]{deb2020advfaces}
Deb, D., Zhang, J., and Jain, A.~K.
\newblock Advfaces: Adversarial face synthesis.
\newblock In \emph{IJCB}, 2020.

\bibitem[Deng et~al.(2020)Deng, Guo, Ververas, Kotsia, and Zafeiriou]{deng2020retinaface}
Deng, J., Guo, J., Ververas, E., Kotsia, I., and Zafeiriou, S.
\newblock Retinaface: Single-shot multi-level face localisation in the wild.
\newblock In \emph{CVPR}, 2020.

\bibitem[Gal et~al.(2023)Gal, Alaluf, Atzmon, Patashnik, Bermano, Chechik, and Cohen-Or]{gal2022image}
Gal, R., Alaluf, Y., Atzmon, Y., Patashnik, O., Bermano, A.~H., Chechik, G., and Cohen-Or, D.
\newblock An image is worth one word: Personalizing text-to-image generation using textual inversion.
\newblock In \emph{ICLR}, 2023.

\bibitem[Goodfellow et~al.(2014)Goodfellow, Shlens, and Szegedy]{goodfellow2014explaining}
Goodfellow, I.~J., Shlens, J., and Szegedy, C.
\newblock Explaining and harnessing adversarial examples.
\newblock \emph{arXiv preprint arXiv:1412.6572}, 2014.

\bibitem[Ho et~al.(2020)Ho, Jain, and Abbeel]{ho2020denoising}
Ho, J., Jain, A., and Abbeel, P.
\newblock Denoising diffusion probabilistic models.
\newblock In \emph{NeurIPS}, 2020.

\bibitem[Hu et~al.(2022)Hu, Wallis, Allen-Zhu, Li, Wang, Wang, Chen, et~al.]{hu2021lora}
Hu, E.~J., Wallis, P., Allen-Zhu, Z., Li, Y., Wang, S., Wang, L., Chen, W., et~al.
\newblock Lora: Low-rank adaptation of large language models.
\newblock In \emph{ICLR}, 2022.

\bibitem[Juefei-Xu et~al.(2022)Juefei-Xu, Wang, Huang, Guo, Ma, and Liu]{juefei2022countering}
Juefei-Xu, F., Wang, R., Huang, Y., Guo, Q., Ma, L., and Liu, Y.
\newblock Countering malicious deepfakes: Survey, battleground, and horizon.
\newblock \emph{IJCV}, 2022.

\bibitem[Kumari et~al.(2023)Kumari, Zhang, Zhang, Shechtman, and Zhu]{kumari2023multi}
Kumari, N., Zhang, B., Zhang, R., Shechtman, E., and Zhu, J.-Y.
\newblock Multi-concept customization of text-to-image diffusion.
\newblock In \emph{CVPR}, 2023.

\bibitem[Li et~al.(2024)Li, Mo, Li, and Wang]{li2024pid}
Li, A., Mo, Y., Li, M., and Wang, Y.
\newblock Pid: prompt-independent data protection against latent diffusion models.
\newblock In \emph{ICML}, 2024.

\bibitem[Li et~al.(2023)Li, Yu, Salem, Backes, Fritz, and Zhang]{li2023unganable}
Li, Z., Yu, N., Salem, A., Backes, M., Fritz, M., and Zhang, Y.
\newblock $\{$UnGANable$\}$: Defending against $\{$GAN-based$\}$ face manipulation.
\newblock In \emph{USENIX Security}, 2023.

\bibitem[Liang et~al.(2023)Liang, Wu, Hua, Zhang, Xue, Song, Xue, Ma, and Guan]{liang2023adversarial}
Liang, C., Wu, X., Hua, Y., Zhang, J., Xue, Y., Song, T., Xue, Z., Ma, R., and Guan, H.
\newblock Adversarial example does good: Preventing painting imitation from diffusion models via adversarial examples.
\newblock In \emph{ICML}, 2023.

\bibitem[Liu et~al.(2023)Liu, Zhong, Zhang, Qin, and Deng]{liu2023enhancing}
Liu, X., Zhong, Y., Zhang, Y., Qin, L., and Deng, W.
\newblock Enhancing generalization of universal adversarial perturbation through gradient aggregation.
\newblock In \emph{ICCV}, 2023.

\bibitem[Liu et~al.(2025)Liu, Zhong, Cui, Zhang, Li, and Deng]{liu2025advcloak}
Liu, X., Zhong, Y., Cui, X., Zhang, Y., Li, P., and Deng, W.
\newblock Advcloak: Customized adversarial cloak for privacy protection.
\newblock \emph{Pattern Recognition}, 2025.

\bibitem[Liu et~al.(2024)Liu, Fan, Dai, Chen, Zhou, and Sun]{liu2024metacloak}
Liu, Y., Fan, C., Dai, Y., Chen, X., Zhou, P., and Sun, L.
\newblock Metacloak: Preventing unauthorized subject-driven text-to-image diffusion-based synthesis via meta-learning.
\newblock In \emph{CVPR}, 2024.

\bibitem[Liu et~al.(2015)Liu, Luo, Wang, and Tang]{liu2015deep}
Liu, Z., Luo, P., Wang, X., and Tang, X.
\newblock Deep learning face attributes in the wild.
\newblock In \emph{ICCV}, 2015.

\bibitem[Madry(2017)]{madry2017towards}
Madry, A.
\newblock Towards deep learning models resistant to adversarial attacks.
\newblock \emph{arXiv preprint arXiv:1706.06083}, 2017.

\bibitem[Mittal et~al.(2012)Mittal, Moorthy, and Bovik]{mittal2012no}
Mittal, A., Moorthy, A.~K., and Bovik, A.~C.
\newblock No-reference image quality assessment in the spatial domain.
\newblock \emph{IEEE TIP}, 2012.

\bibitem[Moosavi-Dezfooli et~al.(2016)Moosavi-Dezfooli, Fawzi, and Frossard]{moosavi2016deepfool}
Moosavi-Dezfooli, S.-M., Fawzi, A., and Frossard, P.
\newblock Deepfool: a simple and accurate method to fool deep neural networks.
\newblock In \emph{CVPR}, 2016.

\bibitem[Moosavi-Dezfooli et~al.(2017)Moosavi-Dezfooli, Fawzi, Fawzi, and Frossard]{moosavi2017universal}
Moosavi-Dezfooli, S.-M., Fawzi, A., Fawzi, O., and Frossard, P.
\newblock Universal adversarial perturbations.
\newblock In \emph{CVPR}, 2017.

\bibitem[Nichol et~al.(2022)Nichol, Dhariwal, Ramesh, Shyam, Mishkin, McGrew, Sutskever, and Chen]{nichol2021glide}
Nichol, A., Dhariwal, P., Ramesh, A., Shyam, P., Mishkin, P., McGrew, B., Sutskever, I., and Chen, M.
\newblock Glide: Towards photorealistic image generation and editing with text-guided diffusion models.
\newblock In \emph{ICML}, 2022.

\bibitem[Poursaeed et~al.(2018)Poursaeed, Katsman, Gao, and Belongie]{poursaeed2018generative}
Poursaeed, O., Katsman, I., Gao, B., and Belongie, S.
\newblock Generative adversarial perturbations.
\newblock In \emph{CVPR}, 2018.

\bibitem[Rombach et~al.(2022)Rombach, Blattmann, Lorenz, Esser, and Ommer]{rombach2022high}
Rombach, R., Blattmann, A., Lorenz, D., Esser, P., and Ommer, B.
\newblock High-resolution image synthesis with latent diffusion models.
\newblock In \emph{CVPR}, 2022.

\bibitem[Ruiz et~al.(2020)Ruiz, Bargal, and Sclaroff]{ruiz2020disrupting}
Ruiz, N., Bargal, S.~A., and Sclaroff, S.
\newblock Disrupting deepfakes: Adversarial attacks against conditional image translation networks and facial manipulation systems.
\newblock In \emph{ECCVW}, 2020.

\bibitem[Ruiz et~al.(2023)Ruiz, Li, Jampani, Pritch, Rubinstein, and Aberman]{ruiz2023dreambooth}
Ruiz, N., Li, Y., Jampani, V., Pritch, Y., Rubinstein, M., and Aberman, K.
\newblock Dreambooth: Fine tuning text-to-image diffusion models for subject-driven generation.
\newblock In \emph{CVPR}, 2023.

\bibitem[Shan et~al.(2020)Shan, Wenger, Zhang, Li, Zheng, and Zhao]{shan2020fawkes}
Shan, S., Wenger, E., Zhang, J., Li, H., Zheng, H., and Zhao, B.~Y.
\newblock Fawkes: Protecting privacy against unauthorized deep learning models.
\newblock In \emph{USENIX Security}, 2020.

\bibitem[Shi et~al.(2024)Shi, Xiong, Lin, and Jung]{shi2024instantbooth}
Shi, J., Xiong, W., Lin, Z., and Jung, H.~J.
\newblock Instantbooth: Personalized text-to-image generation without test-time finetuning.
\newblock In \emph{CVPR}, 2024.

\bibitem[Sohl-Dickstein et~al.(2015)Sohl-Dickstein, Weiss, Maheswaranathan, and Ganguli]{sohl2015deep}
Sohl-Dickstein, J., Weiss, E., Maheswaranathan, N., and Ganguli, S.
\newblock Deep unsupervised learning using nonequilibrium thermodynamics.
\newblock In \emph{ICML}, 2015.

\bibitem[Song et~al.(2021)Song, Meng, and Ermon]{song2020denoising}
Song, J., Meng, C., and Ermon, S.
\newblock Denoising diffusion implicit models.
\newblock In \emph{ICLR}, 2021.

\bibitem[Szegedy(2013)]{szegedy2013intriguing}
Szegedy, C.
\newblock Intriguing properties of neural networks.
\newblock \emph{arXiv preprint arXiv:1312.6199}, 2013.

\bibitem[Terhorst et~al.(2020)Terhorst, Kolf, Damer, Kirchbuchner, and Kuijper]{terhorst2020ser}
Terhorst, P., Kolf, J.~N., Damer, N., Kirchbuchner, F., and Kuijper, A.
\newblock Ser-fiq: Unsupervised estimation of face image quality based on stochastic embedding robustness.
\newblock In \emph{CVPR}, 2020.

\bibitem[Van~Le et~al.(2023)Van~Le, Phung, Nguyen, Dao, Tran, and Tran]{van2023anti}
Van~Le, T., Phung, H., Nguyen, T.~H., Dao, Q., Tran, N.~N., and Tran, A.
\newblock Anti-dreambooth: Protecting users from personalized text-to-image synthesis.
\newblock In \emph{ICCV}, 2023.

\bibitem[Vyas et~al.(2023)Vyas, Kakade, and Barak]{vyas2023provable}
Vyas, N., Kakade, S.~M., and Barak, B.
\newblock On provable copyright protection for generative models.
\newblock In \emph{ICML}, 2023.

\bibitem[Wan et~al.(2024)Wan, He, Song, and Gong]{wan2024prompt}
Wan, C., He, Y., Song, X., and Gong, Y.
\newblock Prompt-agnostic adversarial perturbation for customized diffusion models.
\newblock In \emph{NeurIPS}, 2024.

\bibitem[Wang et~al.(2024{\natexlab{a}})Wang, Tan, Wei, Wu, and Huang]{wang2024simac}
Wang, F., Tan, Z., Wei, T., Wu, Y., and Huang, Q.
\newblock Simac: A simple anti-customization method for protecting face privacy against text-to-image synthesis of diffusion models.
\newblock In \emph{CVPR}, 2024{\natexlab{a}}.

\bibitem[Wang et~al.(2022{\natexlab{a}})Wang, Huang, Chen, Liu, Chen, and Wang]{wang2022anti}
Wang, R., Huang, Z., Chen, Z., Liu, L., Chen, J., and Wang, L.
\newblock Anti-forgery: Towards a stealthy and robust deepfake disruption attack via adversarial perceptual-aware perturbations.
\newblock In \emph{IJCAI}, 2022{\natexlab{a}}.

\bibitem[Wang et~al.(2024{\natexlab{b}})Wang, Guo, Liu, Li, Zhao, Tang, Hu, Tang, and Li]{wang2024stablegarment}
Wang, R., Guo, H., Liu, J., Li, H., Zhao, H., Tang, X., Hu, Y., Tang, H., and Li, P.
\newblock Stablegarment: Garment-centric generation via stable diffusion.
\newblock \emph{arXiv preprint arXiv:2403.10783}, 2024{\natexlab{b}}.

\bibitem[Wang et~al.(2022{\natexlab{b}})Wang, Huang, Ma, Nepal, and Xu]{wang2022deepfake}
Wang, X., Huang, J., Ma, S., Nepal, S., and Xu, C.
\newblock Deepfake disrupter: The detector of deepfake is my friend.
\newblock In \emph{CVPR}, 2022{\natexlab{b}}.

\bibitem[Xiao et~al.(2018)Xiao, Li, Zhu, He, Liu, and Song]{xiao2018generating}
Xiao, C., Li, B., Zhu, J.-Y., He, W., Liu, M., and Song, D.
\newblock Generating adversarial examples with adversarial networks.
\newblock In \emph{IJCAI}, 2018.

\bibitem[Xiong et~al.(2023)Xiong, Wu, Yu, and Zheng]{xiong2023black}
Xiong, L., Wu, Y., Yu, P., and Zheng, Y.
\newblock A black-box reversible adversarial example for authorizable recognition to shared images.
\newblock \emph{Pattern Recognition}, 2023.

\bibitem[Xue et~al.(2023)Xue, Liang, Wu, and Chen]{xue2023toward}
Xue, H., Liang, C., Wu, X., and Chen, Y.
\newblock Toward effective protection against diffusion-based mimicry through score distillation.
\newblock In \emph{ICLR}, 2023.

\bibitem[Yang et~al.(2021)Yang, Dong, Pang, Su, Zhu, Chen, and Xue]{yang2021towards}
Yang, X., Dong, Y., Pang, T., Su, H., Zhu, J., Chen, Y., and Xue, H.
\newblock Towards face encryption by generating adversarial identity masks.
\newblock In \emph{ICCV}, 2021.

\bibitem[Zhang et~al.(2025)Zhang, Jia, Chen, Chen, Zhang, Liu, Ding, and Liu]{zhang2025generate}
Zhang, Y., Jia, J., Chen, X., Chen, A., Zhang, Y., Liu, J., Ding, K., and Liu, S.
\newblock To generate or not? safety-driven unlearned diffusion models are still easy to generate unsafe images... for now.
\newblock In \emph{ECCV}, 2025.

\bibitem[Zhong \& Deng(2022)Zhong and Deng]{zhong2022opom}
Zhong, Y. and Deng, W.
\newblock Opom: Customized invisible cloak towards face privacy protection.
\newblock \emph{IEEE TPAMI}, 2022.

\end{thebibliography}
\bibliographystyle{icml2025}

\newpage
\appendix
\onecolumn
\section{Additional Background}
\label{sec:bg}
\textbf{Diffusion Models}.
Diffusion models \citep{sohl2015deep,ho2020denoising} are a type of generative models that learns the data
distribution via two opposing procedures: a forward pass and a backward pass.
Given an input image $\mathbf{x}_0 \sim q(\mathbf{x})$, the forward process gradually corrupts the data over $T$ timesteps by adding Gaussian noise:
\begin{equation}
    \mathbf{x}_t = \sqrt{\alpha_t} \mathbf{x}_0 + \sqrt{1 - \alpha_t} \boldsymbol{\epsilon}, \quad \boldsymbol{\epsilon} \sim \mathcal{N}(\mathbf{0}, \mathbf{I}),
\end{equation}
where $\{\alpha_t\}_{t=1}^T$ follows a predefined variance schedule controlling noise levels at each timestep $t \in [1, T]$.
The reverse process reconstructs $\mathbf{x}_0$ from $\mathbf{x}_T$ by iteratively predicting and removing noise. A parameterized network $\boldsymbol{\epsilon}_\theta(\mathbf{x}_t, t)$ is used to estimate the noise $\boldsymbol{\epsilon}$ added at timestep $t$. 
The training loss is commonly defined as the $\ell_2$ distance between predicted and actual noise:
\begin{equation}
    \mathcal{L}(\theta, \mathbf{x}_0) = \mathbb{E}_{\mathbf{x}_0, t, \boldsymbol{\epsilon} \sim \mathcal{N}(0,\mathbf{I})} \left\| \boldsymbol{\epsilon} - \boldsymbol{\epsilon}_\theta(\mathbf{x}_t, t) \right\|_2^2,
\end{equation}
where $t$ is uniformly sampled from $\{1, \dots, T\}$.

Text-to-image diffusion models incorporates an additional conditioning signal $c$ (e.g., text prompts) into the noise prediction network:
\begin{equation}
    \mathcal{L}(\theta, \mathbf{x}_0, c) = \mathbb{E}_{\mathbf{x}_0, t, \boldsymbol{\epsilon} \sim \mathcal{N}(0,\mathbf{I})} \left\| \boldsymbol{\epsilon} - \boldsymbol{\epsilon}_\theta(\mathbf{x}_t, t, c) \right\|_2^2.
\end{equation}
Sampling from a diffusion model is an iterative reverse process that progressively denoises the data. 
Denoising Diffusion Implicit Model (DDIM) \cite{song2020denoising} is one of the denoising approaches with a deterministic process:
Following the sampling process of DDIM, the denoising step at $t$ is formulated as:
\begin{equation}
    \mathbf{x}_{t-1} = \sqrt{\alpha_{t-1}} \underbrace{\left( \frac{\mathbf{x}_t - \sqrt{1-\alpha_t} \boldsymbol{\epsilon}_\theta(\mathbf{x}_t, t)}{\sqrt{\alpha_t}} \right)}_{\text{Predicted } \mathbf{x}_0} + \sqrt{1-\alpha_{t-1}-\sigma_t^2} \boldsymbol{\epsilon}_\theta(\mathbf{x}_t, t) + \sigma_t \boldsymbol{\epsilon}_t,
\end{equation}
where $\boldsymbol{\epsilon}_t \sim \mathcal{N}(\mathbf{0}, \mathbf{I})$. 
In our work, we utilize the denoising diffusion implicit model (DDIM) to predict the clean data point.

Recently, \textbf{Latent Diffusion Models (LDMs)}~\cite{rombach2022high} have introduced a novel paradigm by operating in the latent space rather than directly in the high-dimensional data space. Specifically, the source latent variable \( z_0 \) is obtained by encoding a sample \( x_0 \) using an encoder \( \mathcal{E} \), such that \( z_0 = \mathcal{E}(x_0) \). This latent representation can then be reversed to reconstruct the original output through a decoder \( \mathcal{D} \). By conducting the diffusion process in a lower-dimensional latent space, LDMs significantly reduce the computational burden while maintaining the quality of the generated images, making it a promising method for high-resolution image synthesis. The training of latent diffusion models involves a denoising process in the latent space, which is optimized through the following objective function:
\begin{equation}
\mathcal{L}_{LDM}(\theta, \mathbf{x}_0, c) = \mathbb{E}_{z_0={\mathcal{E}(\mathbf{x}_0)},\epsilon,t}[\|\epsilon - \epsilon_\theta(z_t,t, c)\|_2^2]
\end{equation}

\section{Derivation of the Optimization Objective for the Anchor Points $\{c_i\}_{i=1}^N$} \label{sec:derivation}
Here, we provide a detailed derivation from Eq.~\eqref{eq:context_diverse_vanilla} to Eq.~\eqref{eq:context_diverse_after}.
Given a set of images \( \{x_i\}_{i=1}^N \) and an identity condition \( c_{ID} \), our goal is to learn a corresponding text conditions \( c_i \) for each image \( x_i \) such that \( c_i \) best describes \( x_i \). Formally, this can be expressed as:
\begin{equation}
    c_i = \arg\max_c p(c | x_i, c_{ID}). 
\end{equation}
Directly optimizing the posterior probability \( p(c | x_i, c_{ID}) \) is intractable. Following \citet{wan2024prompt}, using Bayes' theorem, we decompose it as:
\begin{equation}
    p(c|x_i, c_{ID}) = \frac{p(x_i, c_{ID}|c) \cdot p(c)}{p(x_i,c_{ID})}. 
\end{equation}
Here, the denominator \( p(x_i, c_{ID}) \) acts as a normalization constant \( Z \), since \( x_i \) and \( c_{ID} \) are conditionally independent given \( c \). Thus, the posterior probability is proportional to:
\begin{equation}
    p(c|x_i, c_{ID}) \propto p(x_i|c, c_{ID}) \cdot p(c). 
\end{equation}
We assume a uniform prior over \( c \), i.e., \( p(c) = \frac{1}{K} \). Under this assumption, the prior term becomes a constant, and the optimization objective simplifies to:
\begin{equation}
    c_i = \arg\max_c p(x_i|c, c_{ID}). 
\end{equation}
Diffusion models maximize the likelihood indirectly by minimizing the noise prediction error. Let \( \epsilon_\theta \) denote the denoising network, \( \epsilon \) the true noise, \( x_{i,t} \) the noisy version of \( x_i \) at diffusion step \( t \), and \( \theta^* \) the frozen parameters from the identity learning stage. Following the diffusion training objective~\cite{ho2020denoising}, maximizing \( p(x_i|c, c_{ID}) \) is equivalent to minimizing:
\begin{equation}
    \min_{c_i} \mathbb{E}_{\epsilon,t} \left\| \epsilon - \epsilon_{\theta^*}\left(x_{i,t}, t, c_i\right) \right\|_2^2. 
\end{equation}
Extending this to all images \( \{x_i\}_{i=1}^N \), the final optimization objective becomes:
\begin{equation}
    \min_{\{c_i\}_{i=1}^N} \mathbb{E}_{\epsilon,t } \sum_{i=1}^N \left\| \epsilon - \epsilon_{\theta^*}\left(x_{i,t}, t, c_i\right) \right\|_2^2
\end{equation}
Given that \( c_{ID} \) serves as the point representing the individual's core identity information, which is typically expected to be highly correlated with the content of the image, we assume that \( c_{ID} \) and \( c_i \) are very close in the textual space. Therefore, the iterative solution for \( c_i \) can be initialized from \( c_{ID} \). This initialization provides a strong starting point for optimizing Eq.~\eqref{eq:context_diverse_after}, ensuring faster convergence and better alignment with the image content.

\section{Cloak Updating Strategy} \label{sec:gradient_update}

When updating the cloak, the Projected Gradient Descent (PGD) technique \cite{madry2017towards} is commonly employed, as follows:
\begin{equation}
\delta_{i} = \text{clip}^\eta_\delta(\delta_{i-1} + \alpha\cdot\text{sign}(\nabla_{\delta} \mathcal{L}(x_t,\delta))),
\end{equation}
where the clip operation constrains the pixel values of $\delta$ within an $\eta$-ball around the original values, and $\mathcal{L}$ refers to the optimization objective defined in Eq.~(\ref{eq:final_obj}).

However, in our practice, directly applying PGD to optimize the cloak can lead to suboptimal results due to gradient instability. During each optimization iteration, a small batch of latents \(x_t\) is sampled from Gaussian noise \(x_T\). The significant variations between minibatches of \(x_t\) sampled in different iterations cause instability in the gradients backpropagated through the loss function. Additionally, the use of the sign operation within the PGD framework introduces quantization errors, further exacerbating the instability caused by gradient fluctuations. As a result, the update directions become inconsistent, undermining the optimization process.

To address this issue, we adopt a strategy similar to \citet{liu2023enhancing} and aggregate multiple small-batch gradients to update the cloak, thereby alleviating the aforementioned problem. Specifically, our strategy involves an inner-outer iteration framework. In the inner loop, we perform multiple optimization iterations, each involving sampling a minibatch of \(x_t\) and computing the loss to obtain diversified gradients. Following \citet{liu2023enhancing}, we also introduce a pre-search step within each inner iteration, where a surrogate cloak \(\delta^{\text{inner}}\) of \(\delta\) is preliminarily updated. This step has been shown to enhance the generalization ability of universal cloaks. In each outer loop iteration, we aggregate the gradients collected from the inner loop to obtain a stable and reliable gradient, which is then used to update the universal cloak.
By accumulating gradient information over multiple rounds, the aggregated gradient estimates become more accurate with reduced variance. This suppresses gradient instability and makes the optimization process of the universal cloak more effective. The complete optimization process with stochastic gradient aggregation is outlined in Algorithm.~\ref{alg:sga}.

We also conduct an ablation study on the gradient updating strategy by replacing it with naive PGD. The results, as shown in Table~\ref{tab:ablation_grad}, demonstrate that the gradient aggregation strategy has a significant impact on the effectiveness of the generated identity-specific cloak.

\begin{algorithm}[]
    \caption{Optimizing Identity-specific Universal Cloak with Stochastic Gradient Aggregation}
    \label{alg:sga}
\begin{algorithmic}[1] 
\REQUIRE Customized diffusion model \( \theta^* \), individual subspace \( Q(c) \), training iterations \( N \), inner iterations for gradient aggregation \( M \), noise budget \( \eta \), PGD step size \( \alpha \)

\STATE {\bfseries Initialize:} \( \delta= 0 \)
\FOR{$n = 1$ {\bfseries to} $N$} 
    \STATE \( \delta_1^{\text{inner}} \gets \delta \)
    \STATE \( g^{\text{Aggs}} = 0 \)
    \FOR{$m = 1$ {\bfseries to} $M$} 
        \STATE Sample \( c \sim Q, t \in U(0, T) \)
        \STATE Sample \( x_t = \texttt{sample}(\theta^*, t, c) \)
        \STATE Obtain \( \hat{x}_t' = \texttt{applyCloak}(x_t, \delta) \)
        \STATE Compute grad. \( g_x = \nabla_{\hat{x}_t'}\| \epsilon_\theta^*(x_t, t, c) - \epsilon_\theta^*(\hat{x}_t', t, c) \|_2^2 \) 
        \STATE \( \delta_{m+1}^{\text{inner}} \gets \text{clip}^\eta_\delta(\delta_{m}^{\text{inner}} + \alpha \cdot \text{sgn}(g_x)) \)
        \STATE \( g^{\text{Aggs}} \gets g^{\text{Aggs}} + g_x \) 
    \ENDFOR
    \STATE \( \delta \gets \text{clip}^\eta_\delta(\delta + \alpha \cdot \text{sgn}(g^{\text{Aggs}})) \) 
\ENDFOR
\STATE {\bfseries Return:} Universal cloak \( \delta^* \) for identity \( k \)
\end{algorithmic}
\end{algorithm}

\begin{table}[ht]
    \centering
    \caption{Ablation study on the gradient updating strategy.}
    \renewcommand{\arraystretch}{1.2}
    \setlength{\tabcolsep}{5pt}
    \begin{tabular}{ccccc}
    \toprule
    Method & BRISQUE$\uparrow$& ISM$\downarrow$& FDFR$\uparrow$ & SER-FIQ$\downarrow$ \\
    \midrule
    w/o gradient aggregation & 26.62 & 0.378 & 0.109 & 0.554 \\
    w/ gradient aggregation & 31.05 & 0.364 & 0.210 & 0.501 \\
    \bottomrule
\end{tabular}
\label{tab:ablation_grad}
\end{table}   

\section{Additional Qualitative Comparisons}
\begin{figure}
    \centering
    \includegraphics[width=0.95\linewidth]{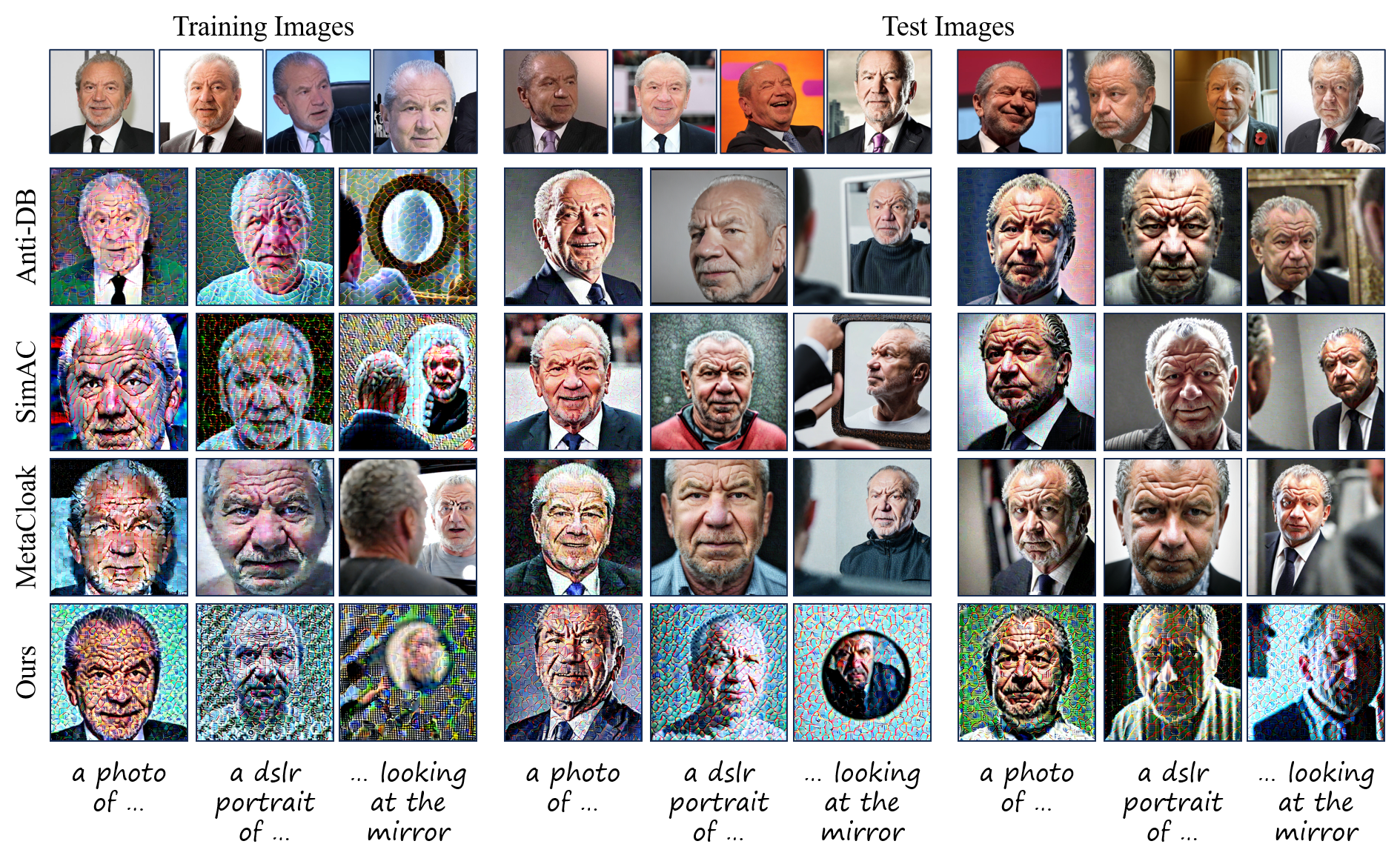}
    \vspace{-6mm}
    \caption{Additional qualitative results on VGGFace2 dataset. The cloaks are generated from the images of training set, then applied on the same training set and different test sets respectively. Each row represents a method, and each column represents a different test prompt.}
    \label{fig:complementary_3}
\end{figure}

\begin{figure}
    \centering
    \includegraphics[width=0.95\linewidth]{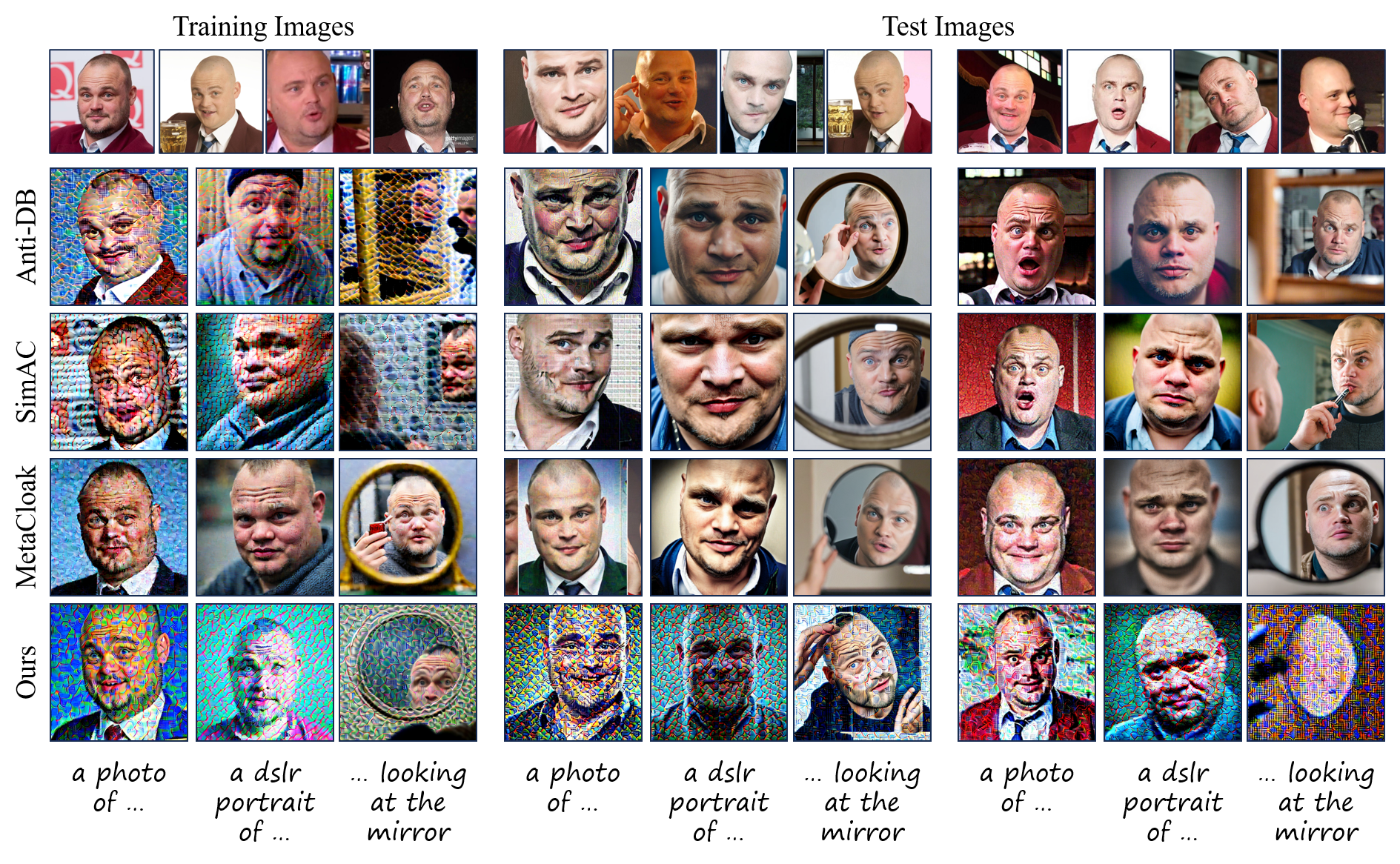}
    \vspace{-6mm}
    \caption{Additional qualitative results on VGGFace2 dataset. The cloaks are generated from the images of training set, then applied on the same training set and different test sets respectively. Each row represents a method, and each column represents a different test prompt.}
    \label{fig:complementary_4}
\end{figure}

\begin{figure}
    \centering
    \includegraphics[width=0.95\linewidth]{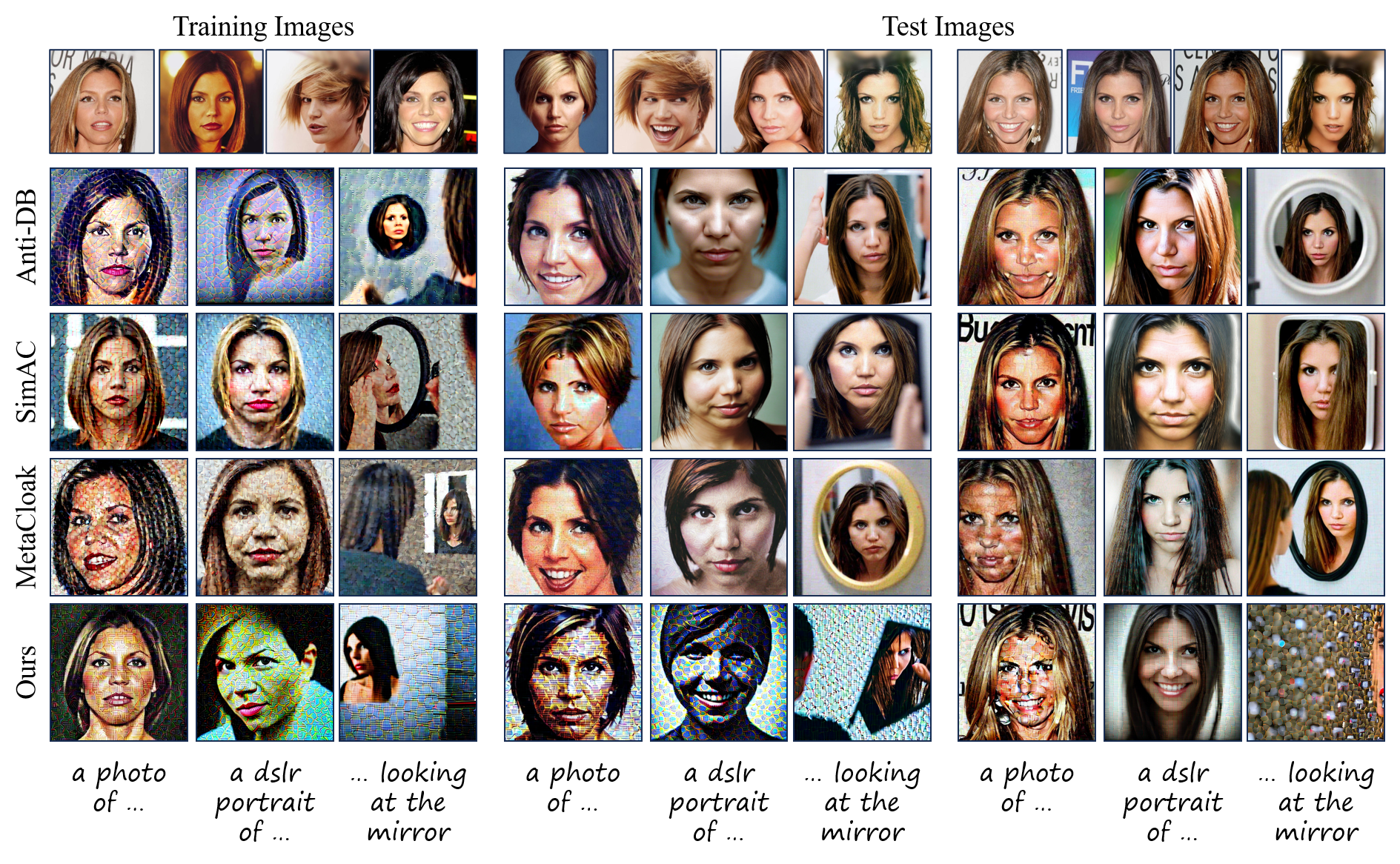}
    \vspace{-6mm}
    \caption{Additional qualitative results on CelebA-HQ dataset. The cloaks are generated from the images of training set, then applied on the same training set and different test sets respectively. Each row represents a method, and each column represents a different test prompt.}
    \label{fig:complementary_1}
\end{figure}

\begin{figure}
    \centering
    \includegraphics[width=0.95\linewidth]{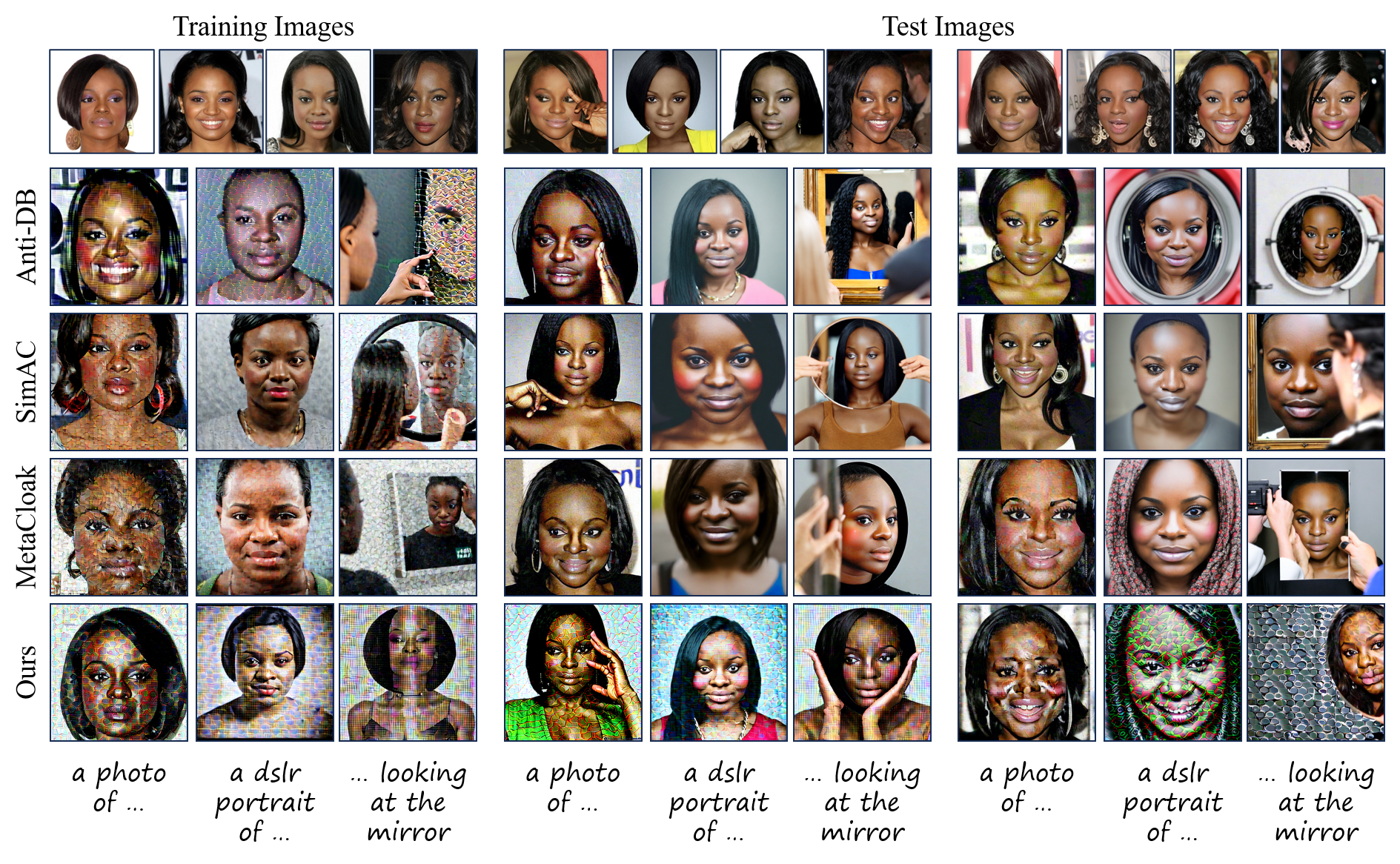}
    \vspace{-6mm}
    \caption{Additional qualitative results on CelebA-HQ dataset. The cloaks are generated from the images of training set, then applied on the same training set and different test sets respectively. Each row represents a method, and each column represents a different test prompt.}
    \label{fig:complementary_2}
\end{figure}

Additional comparison results from our main experiment are presented in 
Figure~\ref{fig:complementary_3} and \ref{fig:complementary_4} for the VGGFace2 dataset, and in Figure~\ref{fig:complementary_1} and \ref{fig:complementary_2} for the CelebA-HQ dataset. 
These results, in conjunction with those presented in Figure~\ref{fig:main_result} of the main paper, substantiate the effectiveness of our method in generating identity-specific cloaks that robustly protect facial privacy and demonstrate its strong generalization capability across all face images of an individual.

\end{document}